\documentclass[conference]{IEEEtran}
\IEEEoverridecommandlockouts
\usepackage{cite}
\usepackage{amsmath,amssymb,amsfonts}
\usepackage{algorithmic}
\usepackage{graphicx}
\usepackage{textcomp}
\usepackage{xcolor}
\usepackage{epsfig}
\usepackage{booktabs}
\usepackage{graphicx}
\usepackage{multirow}
\usepackage{amsmath}
\usepackage{amssymb}
\usepackage{colortbl}
\makeatletter
\newcommand{\ssymbol}[1]{\@fnsymbol{#1}}
\makeatother
\usepackage{float} 
\usepackage{pifont}
\usepackage[misc]{ifsym}
\let\OLDthebibliography\thebibliography
\renewcommand\thebibliography[1]{
	\OLDthebibliography{#1}
	\setlength{\parskip}{0pt}
	\setlength{\itemsep}{0pt plus 0.3ex}
}
\usepackage[flushmargin]{footmisc}
\renewcommand{\footnoterule}{%
    \kern -3pt
    \hrule width 0.3\textwidth height 0.4pt
    \kern 2pt
}
\usepackage{xcolor}
\usepackage{svg} % 需使用包
\definecolor{mycustompurple}{RGB}{154, 36, 79} % 定义自己的颜色
\usepackage[utf8]{inputenc}
\usepackage{hyperref} % 引入超链接宏包

\hypersetup{
	colorlinks=false,            % 激活链接颜色，去掉链接边框
	linkcolor=red,              % 文档内部链接颜色（如图表等引用）
	citecolor=green,            % 文献引用链接颜色
	filecolor=mycustompurple,   % 文件链接颜色
	urlcolor=magenta            % 外部URL链接颜色
}

\def\BibTeX{{\rm B\kern-.05em{\sc i\kern-.025em b}\kern-.08em
    T\kern-.1667em\lower.7ex\hbox{E}\kern-.125emX}}
\begin{document}

\title{AdaMHF: Adaptive Multimodal Hierarchical Fusion for Survival Prediction}

\author{
    \textbf{Shuaiyu Zhang}$^{1,2\ssymbol{2}}$ \quad \textbf{Xun Lin}$^{1,3\ssymbol{2}}$ \quad \textbf{Rongxiang Zhang}$^2$ \quad \textbf{Yu Bai}$^{1\ssymbol{1}}$ \\ 
    \textbf{Yong Xu}$^5$ \quad \textbf{Tao Tan}$^3$ \quad \textbf{Xubin Zheng}$^1$ \quad \textbf{Zitong Yu}$^{1,4\ssymbol{1}}$
    \thanks{
        \quad This work was supported by Guangdong Basic and Applied Basic Research Foundation (Grant No. 2023A1515140037), Guangdong Basic and Applied Basic Research Foundation (Grant No. 2023B0303000010), and Guangdong Research Team for Communication and Sensing Integrated with Intelligent Computing (Project No. 2024KCXTD047). The computational resources are supported by SongShan Lake HPC Center (SSL-HPC) in Great Bay University.
        
        $^\ssymbol{1}$Corresponding authors.

        $^\ssymbol{2}$These authors contribute equally.
    } \\
    $^1$  School of Computing and Information Technology, Great Bay University \\
    $^2$ Harbin Institute of Technology \quad
    $^3$ Beihang University \quad
    $^4$ Macao Polytechnic University \\
    $^5$ Dongguan Key Laboratory for Intelligence and Information Technology \\
    $^6$ Shenzhen Key Laboratory of Visual Object Detection and Recognition, Harbin Institute of Technology Shenzhen
}

\maketitle

\begin{abstract}
The integration of pathologic images and genomic data for survival analysis has gained increasing attention with advances in multimodal learning. However, current methods often ignore biological characteristics, such as heterogeneity and sparsity, both within and across modalities, ultimately limiting their adaptability to clinical practice.
 To address these challenges, we propose AdaMHF: \textbf{A}daptive \textbf{M}ultimodal \textbf{H}ierarchical \textbf{F}usion, a framework designed for efficient, comprehensive, and tailored feature extraction and fusion. AdaMHF is specifically adapted to the uniqueness of medical data, enabling accurate predictions with minimal resource consumption, even under challenging scenarios with missing modalities. Initially, AdaMHF employs an experts expansion and residual structure to activate specialized experts for extracting heterogeneous and sparse features. Extracted tokens undergo refinement via selection and aggregation, reducing the weight of non-dominant features while preserving comprehensive information.
 Subsequently, the encoded features are hierarchically fused, allowing multi-grained interactions across modalities to be captured. Furthermore, we introduce a survival prediction benchmark designed to resolve scenarios with missing modalities, mirroring real-world clinical conditions. Extensive experiments on TCGA datasets demonstrate that AdaMHF surpasses current state-of-the-art (SOTA) methods, showcasing exceptional performance in both complete and incomplete modality settings. Code is available in \href{https://github.com/Curry30Messi/AdaMHF}{AdaMHF}.
\end{abstract}


\vspace{-0.8em}
\section{Introduction}\label{sec:intro}
\vspace{-0.1em}
% Everyone can use \by{} \lx{} \yzt{} to highlight the modified sections.
Survival prediction evaluates mortality risk based on clinical data, offering critical support for treatment decisions. Recently, deep learning has become central to medical data prediction~\cite{xie2024fusionmamba,clam}. Traditional survival analysis used single-modality data, such as genomics or pathology, with deep learning models tailored to each. In pathology, multi-instance learning (MIL)~\cite{transmil,attentionmil,deepmil} handles whole slide images (WSIs) by treating them as bags to identify predictive regions. For genomics, self-normalizing neural networks (SNNs)~\cite{snn} extract features for survival prediction. Although single-modality prediction has shown promising results, it fails to fully utilize medical data for complementary analysis and alignment. Consequently,  multimodal techniques have gained popularity, combining diagnostic images and genomics to improve accuracy. Transformer-based models~\cite{cmta, li2022hfbsurv, porpoise} focus on cross-modal interactions, utilizing attention mechanisms to capture relationships between different modalities. Furthermore, approaches like~\cite{multi_transport, coattention} emphasize inter-modal similarities, leading to unified representations with stronger correlations.
% Despite these advancements, extracting essential information from complex, heterogeneous cancer data remains challenging. Current methods not only overlook the heterogeneity of clinical features and the computational requirements of large-scale data, failing to effectively integrate global and local features during fusion, but also struggle with the difficulties of obtaining comprehensive clinical data in scenarios where lesions are distant or blood samples are limited. These limitations hinder the applicability of existing methods in practical clinical settings.

\begin{figure*}[t]
	\vspace{-2.2em}
	\centering
	\includegraphics[width=0.86\textwidth]{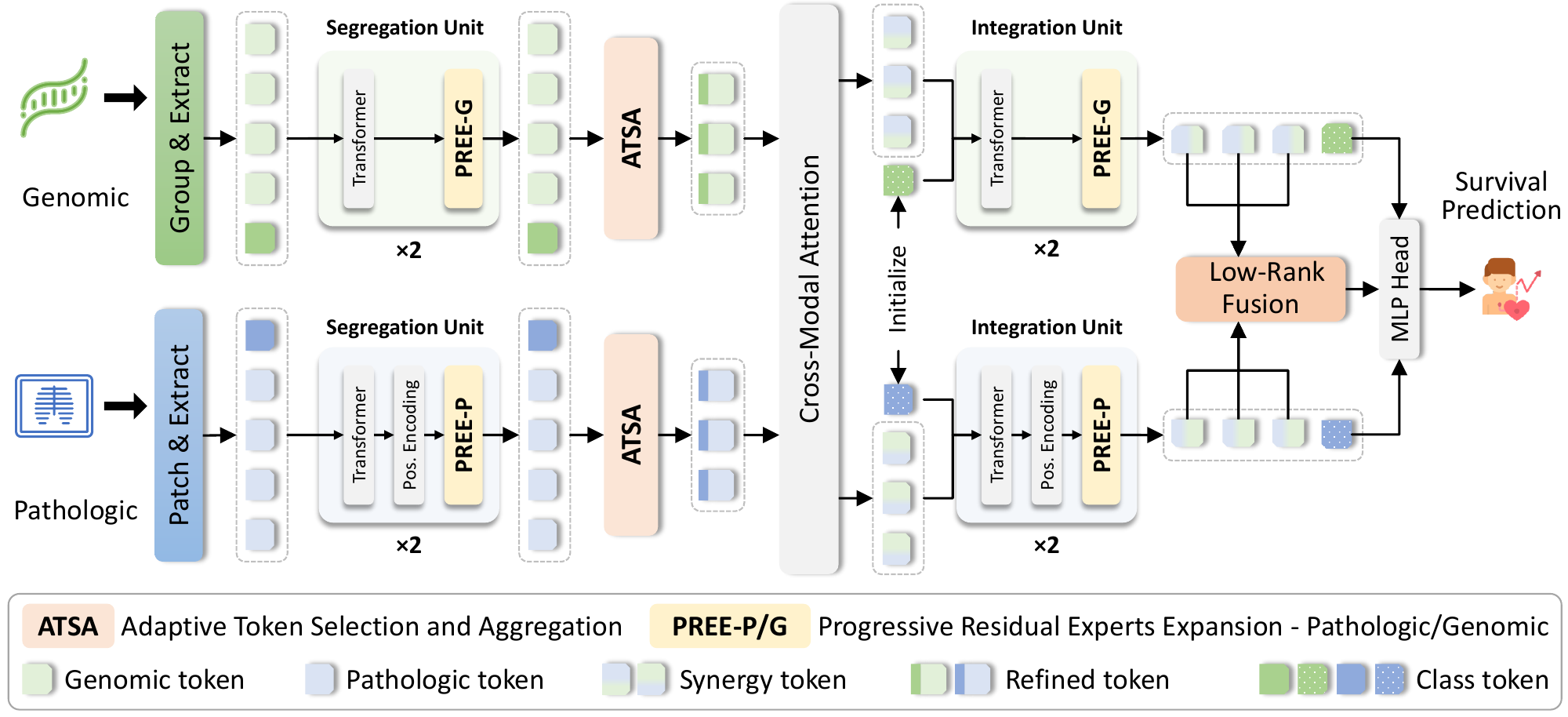}
	\vspace{-1.0em}
	\caption{
		The AdaMHF framework processes preprocessed genomic and pathlogic data through two identical units: the segregation and integration units, both using the PREE module to address heterogeneity. Outputs are combined via a hierarchical fusion mechanism to enhance predictive capability, with interaction facilitated by cross-attention module. Additionally, the ATSA module is specifically designed to ensure model efficiency and resolve sparsity in the data.
	}
	\label{fig:pipline}
	\vspace{-1.2em}
\end{figure*}

% Despite these advancements, extracting essential information from complex and heterogeneous cancer data remains an unresolved challenge due to two unique challenges in real-world scenarios:  
Despite advancements in multimodal fields, extracting meaningful information from cancer data remains an unresolved dilemma due to two indispensable characteristics: \textbf{(1) Sparsity.} Multimodal methods often process high-dimensional genomic data and gigapixel WSIs, where only a small portion is relevant for survival analysis~\cite{coattention}. This \textbf{intra-modal sparsity} increases computational costs and risks capturing spurious associations, such as correlations between survival outcomes and non-pathological regions, thereby degrading clinical prediction performance. Meanwhile, \textbf{inter-modal sparsity}, such as the difficulty of obtaining complete multimodal data due to distant lesions or insufficient blood samples, often leads to missing modalities during inference, which existing methods fail to address. \textbf{(2) Heterogeneity.} Clinical data frequently exhibit \textbf{intra-modal heterogeneity}, influenced by factors such as genetic mutations, and differences in tumor microenvironments. These patient-specific variations complicate feature extraction, and many methods~\cite{transmil,deepmil,zhang2022dtfd} struggle to adapt effectively. Additionally, \textbf{inter-modal heterogeneity}, stemming from the fundamentally different characteristics of pathological images and genomic data~\cite{cmta}, requires specialized processing to appropriately handle each modality.

These limitations hinder the applicability of existing methods in clinical settings. To address these challenges, we propose \textbf{AdaMHF}: an adaptive multimodal hierarchical fusion framework that efficiently combines multi-grained features for comprehensive survival prediction. Our framework consists of two components: a segregation unit and an integration unit. Both units employ the progressive residual expert expansion (PREE) module to manage heterogeneous multimodal data. The expert selection strategy allows for better handling of individual characteristics and provides flexibility in dealing with missing modalities. Furthermore, an adaptive token selection and aggregation (ATSA) module filters redundancy while preserving essential information. Particularly, the router mechanism in ATSA, which selectively retains tokens, significantly reduces the weight of missing modalities, thereby improving robustness when some modalities are unavailable. Then the global tokens and local tokens are hierarchically combined to generate the final prediction, based on the local information obtained via low-rank fusion. This approach reduces computational complexity while maintaining critical information. Hence, the main contributions of our work are as follows:

% \vspace{-0.2em}

\begin{itemize}

	% \vspace{-0.3em}
\item We propose \textbf{AdaMHF}, an adaptive multimodal hierarchical fusion framework that first addresses both sparsity and heterogeneity in bioinformatics for survival analysis.

\item We design the PREE module to activate specialized experts for tailored feature extraction to handle heterogeneity and address missing modality scenarios, while the ATSA mechanism tackles sparsity and improves computational efficiency. During the fusion process, local context is hierarchically integrated with global tokens, ensuring a more comprehensive representation.

	% \vspace{-0.6em}
	\item We establish a missing-modality benchmark and conduct extensive experiments showing that AdaMHF achieves SOTA performance with complete modalities and maintains outstanding results even with incomplete modalities.
	
\end{itemize}   
\vspace{-0.2em}

% \vspace{-1.0em}
\section{Methodology}
% \vspace{-0.3em}
\label{sec:method}

\subsection{Problem Formulation for Survival Prediction}
% \vspace{-0.5em}

\label{sec:survival}
% Following prior work~\cite{transmil,multi_transport}, we define the multimodal data for the $i$-th patient as tissue data $\boldsymbol x_{p}^{(i)}$, sequence data $\boldsymbol x_{g}^{(i)}$, right-censorship status $c_i \in \{0, 1\}$, and overall survival time $t_i \in R^{+}$. Thus, $X_i = (\boldsymbol x_{p}^{(i)}, \boldsymbol x_{g}^{(i)}, c_i, t_i)$.

Following prior works~\cite{li2022hfbsurv,wang2021gpdbn}, whole slide images (WSIs) are divided into patches, and features are extracted using a pre-trained network (e.g., CLAM~\cite{clam}). The omic data is categorized into six groups: 1) Tumor Suppression, 2) Oncogenesis, 3) Protein Kinases, 4) Cellular Differentiation, 5) Transcription, and 6) Cytokines and Growth, with these categories concatenated and processed through a self-normalizing neural network (SNN)~\cite{snn} to extract gene features. This process yields $\boldsymbol x_p \in \mathbb{R}^{n_p \times d}$ for WSIs and $\boldsymbol x_g \in \mathbb{R}^{n_g \times d}$ for genomic data, where $n_p$ denotes the number of patches, and $n_g = 6$ represents the number of genomic categories. Additionally, the overall survival time is defined as $t_i \in \mathbb{R}^{+}$.

AdaMHF takes $\boldsymbol x_p$ and $\boldsymbol x_g$ to obtain the final multimodal feature $\boldsymbol x_n$. The hazard function $f_{h}(t|\boldsymbol x_n) \in [0, 1]$ estimates the probability of a death event at time $t$, and the cumulative survival function is
$f_{s}(t|\boldsymbol x_n) = \prod_{i=1}^{t} (1 - f_{h}(t_i|\boldsymbol x_n))$.
% \begingroup
% \small
% % \vspace{-0.2em}
% \begin{equation}
% $f_{s}(t|\boldsymbol x_n) = \prod_{i=1}^{t} (1 - f_{h}(t_i|\boldsymbol x_n))$.
% \end{equation}
% \endgroup
We regard the negative log-likelihood (NLL) with censorship as the loss function of survival prediction~\cite{cmta,coattention,porpoise}:

% \begingroup
% \small
% \vspace{-0.2em}
\begin{equation}
	\begin{aligned}
		\label{eq:loss-surv}
		\mathcal L_{surv} = & - c \log(f_{s}(t|\boldsymbol x_n)) \\
		& - (1 - c) \log(f_{s}(t-1|\boldsymbol x_n)) \\
		& - (1 - c) \log(f_{h}(t|\boldsymbol x_n)).
	\end{aligned}
	\vspace{-0.2em}
\end{equation}
% \endgroup

% \vspace{-1em}
\subsection{Overall Framework}
\label{sec:adam}
% \vspace{-0.5em}   
To tackle the inter- and intra-modal sparsity and heterogeneity challenges in survival prediction mentioned in Sec.~\ref{sec:intro}, we propose AdaMHF, a novel framework depicted in Fig.~\ref{fig:pipline}. AdaMHF enables to facilitate targeted feature extraction, token refinement and hierarchical fusion, effectively addressing existing challenges and the issue of missing modalities. Initially, AdaMHF processes genomic data ($\boldsymbol{x}_g$) and pathological images ($\boldsymbol{x}_p$) via a segregation unit. This unit comprises a transformer block and modality-specific progressive residual expert expansion (PREE) component (Section~\ref{sec:pree}) applied in two stages to capture both shallow and deep information. For genomic data, we utilize PREE-G, defined as follows:

% \begingroup
% \small

\vspace{-1.3em}
\begin{align}
	\boldsymbol x_g^{(t)} &= \text{Transformer Block}(\boldsymbol x_g^{(t-1)}), \\
	\boldsymbol x_g^{(t+1)} &= \text{PREE-G}(\boldsymbol x_g^{(t)}).
\end{align}
\vspace{-1.5em}
% \endgroup

For pathological slides, after obtaining $\boldsymbol x_p^{(t)}$, we additionally apply the positional embedding and PREE-P to generate $\boldsymbol x_p^{(t+1)}$, with positional encoding implemented via the embedded position encoding generator (EPEG) as per~\cite{cmta}.

\begin{figure*}[t]
	\centering
	\vspace{-2em}
	\includegraphics[width=0.82\textwidth]{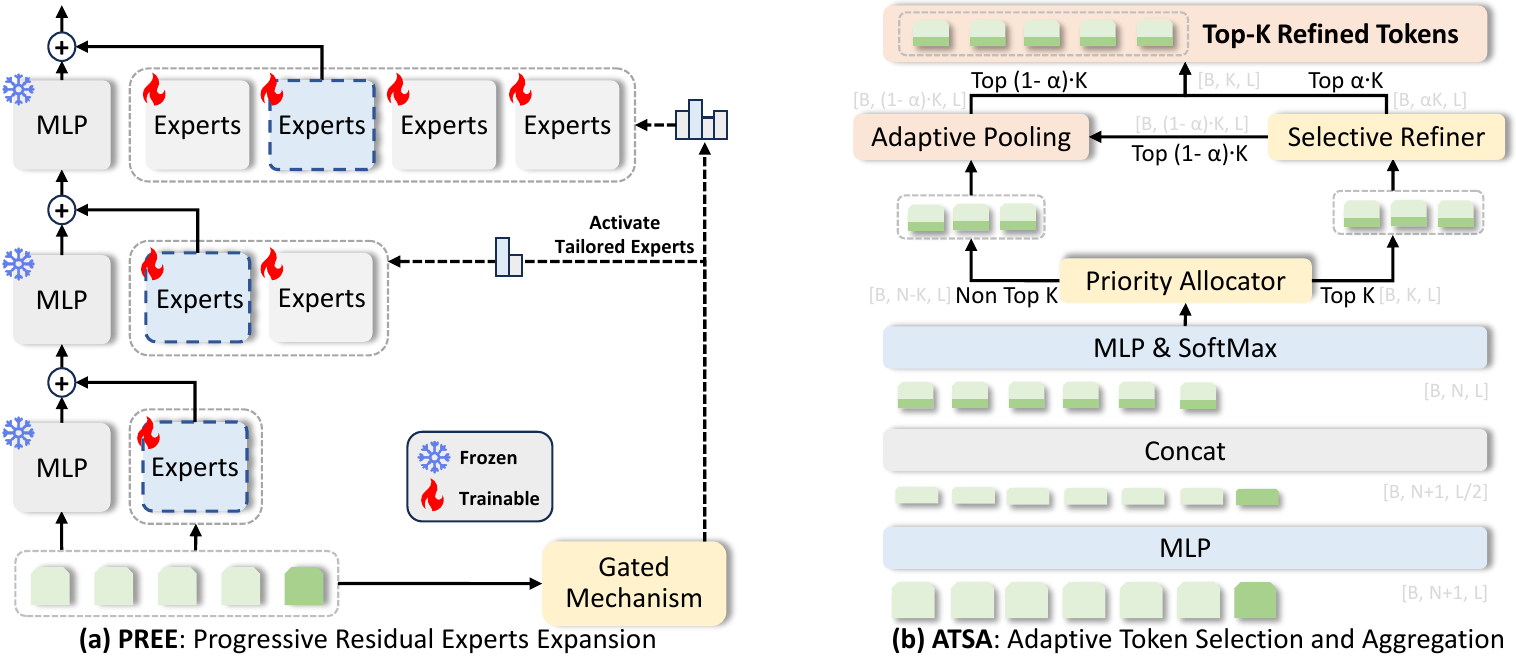}
	\vspace{-0.5em}
	\caption{
The specific structures of the PREE and ATSA modules:
(a) The structure of PREE, where the expert is implemented as CNN for the pathological modality and as SNN for genomic data. (b) The detailed structure of ATSA, where the input consists of a class token and remaining patch-based tokens. Here, $B$ is the batch size, $N$ is the number of tokens, and $L$ is the length of each token.
	}
	\label{fig:details}
	\vspace{-1.5em}
\end{figure*}

Following the segregation unit, the adaptive token selection and aggregation (ATSA) module (Section~\ref{sec:atsa}) reduces tokens while preserving essential information. Inspired by~\cite{cmta}, we employ a cross-modal attention module to facilitate interactions between two modalities. The enriched fused tokens, $x_g^{\star}$ and $x_p^{\star}$, are then sent to the integration unit for deeper interaction:
% \vspace{-0.5em}
% \begingroup
% \small
\begin{align}
	\boldsymbol x_p^{(t+1)} = \text{ATSA}(\boldsymbol x_p^{(t)}), \quad \boldsymbol x_g^{(t+1)} = \text{ATSA}(\boldsymbol x_g^{(t)}), \\
	\boldsymbol x_g^{(t+2)\star}, \boldsymbol x_p^{(t+2)\star} = \text{Cross-Attention}(\boldsymbol x_g^{(t+1)}, \boldsymbol x_p^{(t+1)}).
\end{align}
\vspace{-1.5em}
% \endgroup

The integration unit mirrors the segregation module, ensuring consistent feature extraction and minimizing the risk of introducing conflicting representations to enhance prediction accuracy. Finally, hierarchical fusion is performed by deriving the local output $\tilde{\boldsymbol x}_n^{(t+1)\star}$ from neighborhood tokens via low-rank fusion (Section~\ref{sec:fusion}) and obtaining the global output $\tilde{\boldsymbol x}_n^{cls\star}$ through class token integration. These outputs are then linearly combined for the final prediction:

\vspace{-1em}
% \begingroup
% \small
\begin{align}
	\tilde{\boldsymbol x}_n^{(t+1)\star} &= \text{Low-Rank Fusion}(\boldsymbol x_g^{(t)\star}, \boldsymbol x_p^{(t)\star}), \label{eq:fusion1} \\
	\tilde{\boldsymbol x}_n^{cls\star} &= \text{MLP}(\boldsymbol x_g^{cls\star} \oplus \boldsymbol x_p^{cls\star}), \label{eq:fusion2} \\
	\tilde{\boldsymbol x}_n^{(t+2)\star} &= (1- \beta) \cdot \tilde{\boldsymbol x}_n^{cls\star} + \beta \tilde{\boldsymbol x}_n^{(t+1)\star}, \label{eq:fusion3} \\
	T_1, \ldots, T_t &= \text{sigmoid}\left(\text{MLP}(\tilde{\boldsymbol x}_n^{(t+2)\star})\right),
\end{align}
% \endgroup

where $\beta$ is a hyperparameter and $T_i$ is the probability of cancer recurrence at time $i$. 
For more details about the loss function, preprocessing process, and the mechanism of the cross-attention, please see the supplementary material.

% \begin{figure}[ht]
	% \centering
	
	% \includegraphics[width=0.50\textwidth]{Conference-LaTeX-template_10-17-19/img/details.pdf}
	% \vspace{-1.2em}
	%   \caption{
		% The specific structures of the PREE and ATSA modules: (a) Structure of PREE, where the MLP is fixed and the experts are trainable; (b) Detailed structure of ATSA, with input consisting of a class token and the remaining patch-based tokens.
		%   }
	% \label{fig:details}
	% \vspace{-1.2em}
	% \end{figure}

% \vspace{-0.8em}
\subsection{Progressive Residual Experts Expansion}
\label{sec:pree}
% \vspace{-0.5em}

In this subsection, we introduce the progressive residual experts expansion (PREE) module in both the segregation and integration units to address feature inhomogeneity and incompleteness across modalities. As shown in Fig.~\ref{fig:details}(a), an input feature, such as the pathological image $\boldsymbol x_p^{t}$, traverses a progressively expanding experts network that extracts features stepwise, yielding multi-scale representations from shallow to deep layers. The expert selection mechanism~\cite{tu2024overview,huang2024empirical} enables the PREE to effectively manage individual representations from diverse patients characterized by functionality, addressing intra-modal heterogeneity, reducing computational complexity, and handling missing-modality scenarios by dynamically activating experts tailored to specific modalities.

In the pathological modality, the expert network uses a progressively scaled convolutional neural network (CNN), increasing experts from 1 to 4 across layers. Each layer selects a single expert to adaptively manage complex data structures, aligning with the intricacies of bioinformatics. This approach efficiently balances feature complexity and computational cost and enhances model generalization compared to other settings. Expert selection is governed by a gating mechanism using an MLP and Softmax, selecting the highest-scoring expert at each layer. To preserve shallow features, each layer utilizes a fixed MLP for residual connections with the features extracted by experts. The parameters of this fixed MLP are initialized from the CLAM~\cite{clam} model, ensuring robust extraction of fundamental features. For the genomic modality $\boldsymbol x_g^{t}$, a similar approach is taken, but experts are implemented as self-normalizing neural networks (SNNs). Specilized experts for each modality effectively addresses the inter-modality heterogeneity.
Further details on the MoE architecture, the rationale behind selecting CNNs for pathology, SNNs for genomics, and MLP initialization are provided in the supplementary material.

% \vspace{-1.2em}
\subsection{Adaptive Token Selection and Aggregation}
\label{sec:atsa}
% \vspace{-0.5em}

% \begingroup
% \small
\begin{table*}[t]
    \vspace{-2.0em}
    \centering
    \caption{Performance of AdaMHF on five TCGA public datasets. P and G represent the histological and genomic modalities respectively. The best results are highlighted in \textbf{bold}, and the second-best results are \underline{underlined}.}
    \vspace{-0.5em}
    \label{tab:main} 
    \resizebox{\linewidth}{!}{%
        \scriptsize % Change font size to small here
        \begin{tabular}{lcccccccc} 
            \toprule
            \textbf{Models}         & \textbf{P} & \textbf{G} & \textbf{BLCA}          & \textbf{BRCA}          & \textbf{UCEC}          & \textbf{GBMLGG}        & \textbf{LUAD}          & \textbf{Overall} \\ 
            \midrule
            SNN~\cite{snn}          &            & \checkmark & $0.618 \pm 0.022$      & $0.624 \pm 0.060$      & $0.679 \pm 0.040$      & $0.834 \pm 0.012$      & $0.611 \pm 0.047$      & $0.673$         \\
            SNNTrans~\cite{snn}     &            & \checkmark & $0.645 \pm 0.042$      & $0.647 \pm 0.058$      & $0.632 \pm 0.032$      & $0.828 \pm 0.015$      & $0.633 \pm 0.049$      & $0.677$         \\
            \midrule
            AttnMIL~\cite{attentionmil} & \checkmark &            & $0.599 \pm 0.048$      & $0.609 \pm 0.065$      & $0.658 \pm 0.036$      & $0.818 \pm 0.025$      & $0.620 \pm 0.061$      & $0.661$         \\
            CLAM-MB~\cite{clam}     & \checkmark &            & $0.565 \pm 0.027$      & $0.578 \pm 0.032$      & $0.609 \pm 0.082$      & $0.776 \pm 0.034$      & $0.582 \pm 0.072$      & $0.622$         \\
            CLAM-SB~\cite{clam}     & \checkmark &            & $0.559 \pm 0.034$      & $0.573 \pm 0.044$      & $0.644 \pm 0.061$      & $0.779 \pm 0.031$      & $0.594 \pm 0.063$      & $0.629$         \\
            TransMIL~\cite{transmil}& \checkmark &            & $0.575 \pm 0.034$      & $0.666 \pm 0.029$      & $0.655 \pm 0.046$      & $0.798 \pm 0.043$      & $0.642 \pm 0.046$      & $0.667$         \\
            DeepAttnMISL~\cite{deepmil} & \checkmark &          & $0.504 \pm 0.042$      & $0.524 \pm 0.043$      & $0.597 \pm 0.059$      & $0.734 \pm 0.029$      & $0.548 \pm 0.050$      & $0.581$         \\
            DTFD-MIL~\cite{zhang2022dtfd} & \checkmark &        & $0.546 \pm 0.021$      & $0.609 \pm 0.059$      & $0.656 \pm 0.045$      & $0.792 \pm 0.023$      & $0.585 \pm 0.066$      & $0.638$         \\
            \midrule
            MCAT~\cite{coattention} & \checkmark & \checkmark & $0.672 \pm 0.032$      & $0.659 \pm 0.031$      & $0.649 \pm 0.043$      & $0.835 \pm 0.024$      & $0.659 \pm 0.027$      & $0.695$         \\
            Porpoise~\cite{porpoise}& \checkmark & \checkmark & $0.636 \pm 0.024$      & $0.652 \pm 0.042$      & $0.695 \pm 0.032$      & $0.834 \pm 0.017$      & $0.647 \pm 0.031$      & $0.693$         \\
            MOTCat~\cite{multi_transport} & \checkmark & \checkmark & $0.683 \pm 0.026$ & \underline{$0.673 \pm 0.006$} & $0.675 \pm 0.040$ & $0.849 \pm 0.028$ & $0.670 \pm 0.038$ & $0.710$ \\
            HFBSurv~\cite{li2022hfbsurv} & \checkmark & \checkmark & $0.639 \pm 0.027$ & $0.647 \pm 0.034$ & $0.642 \pm 0.044$ & $0.838 \pm 0.013$ & $0.650 \pm 0.050$ & $0.683$ \\
            GPDBN~\cite{wang2021gpdbn} & \checkmark & \checkmark & $0.635 \pm 0.025$ & $0.654 \pm 0.033$ & $0.683 \pm 0.052$ & \underline{$0.854 \pm 0.024$} & $0.640 \pm 0.047$ & $0.693$ \\
            CMTA~\cite{cmta}        & \checkmark & \checkmark & \underline{$0.691 \pm 0.042$} & $0.667 \pm 0.043$ & \underline{$0.697 \pm 0.040$} & $0.853 \pm 0.011$ & \underline{$0.686 \pm 0.035$} & \underline{$0.719$} \\
            \textbf{AdaMHF (Ours)}  & \checkmark & \checkmark & \textbf{0.708} $\pm$ \textbf{0.027} & \textbf{0.691} $\pm$ \textbf{0.016} & \textbf{0.716} $\pm$ \textbf{0.041} & \textbf{0.865} $\pm$ \textbf{0.009} & \textbf{0.706} $\pm$ \textbf{0.024} & \textbf{0.737} \\
            \bottomrule
        \end{tabular}%
    }
    \vspace{-1.3em}
\end{table*}
	
	% \endgroup

To efficiently handle enormous gene groups and gigapixel WSIs, we introduce the adaptive token selection and aggregation (ATSA) module. This module identifies the most informative tokens while preserving the general information for accurate predictions. In real inference scenarios, tokens from missing modalities are significantly reduced to lower their fusion weight in the final prediction, thereby improving the model's adaptability. As shown in Fig.~\ref{fig:details}(b), the ATSA module selects the TopK informative tokens from the sequence of tokens within a specific modality. For WSIs, the input is represented as $\boldsymbol x_p = \{ \boldsymbol x_p^1, \ldots, \boldsymbol x_p^{n-1}, \boldsymbol x_p^{cls} \} \in \mathbb{R}^{n_p \times d}$, where $\boldsymbol x_p^{cls}$ serves as the holistic pathological feature. Initially, we apply an MLP to $\boldsymbol x_p$ to reduce dimensions and concatenate the global token with each local token. All concatenated features then pass through another MLP and a Softmax layer to predict importance scores. The TopK tokens with the highest scores are selected for further processing, where $K$ is determined by the Priority Allocater, as formulated below:

\vspace{-1.2em}
% \begingroup
% \small
\begin{align}
	\boldsymbol x_p^{i} &= \text{MLP}(\boldsymbol x_p^{i}) \oplus \text{MLP}(\boldsymbol x_p^{cls}), \label{eq:atsa1} \\
	\boldsymbol s_i &= \text{SoftMax}\left(\text{MLP}(\boldsymbol x_p^{i})\right), \label{eq:atsa2} \\
	\boldsymbol x_p^{\text{TopK}} &= \{\boldsymbol x_p^{i} \ | i \in \text{Indices of TopK}(\boldsymbol s)\} \in \mathbb{R}^{k \times d}. \label{eq:atsa3}
\end{align}
% \endgroup

\vspace{-0.5em}
After obtaining $\boldsymbol{x}_p^{\text{TopK}}$, we utilize the Selective Refiner module to assign a ratio $\alpha$, which ensures the preservation of critical tokens without losing essential information. A total of $K \cdot \alpha$ tokens $ \boldsymbol x_p^{\text{Top} K \cdot \alpha}$are retained from these TopK tokens. For the remaining $K \cdot (1-\alpha)$ tokens $ \boldsymbol x_p^{ \text{Top} K \cdot (1-\alpha)}$, including those initially marked for discarding $ \boldsymbol x_p^{\text{Non-TopK}}$, we apply adaptive pooling to create a new informative set of $K \cdot (1-\alpha)$ tokens, denoted as $\boldsymbol{x}_p^{\text{Top}^{\star}K \cdot (1-\alpha)}$. These tokens are then combined with the previously selected $\boldsymbol{x}_p^{\text{Top}K \cdot \alpha}$ to form the final set $\boldsymbol{x}_p^{\text{TopK}}$.
Both the Priority Allocator and Selective Refiner employ a simple router mechanism for their implementation. This strategy preserves information from the entire WSI while maintaining the most critical tokens without disruption, which can be summarized by the following equations:

\vspace{-1em}
% \begingroup
% \small
\begin{align}
	\boldsymbol x_p^{\text{Top}^{\star}K \cdot (1-\alpha)} &= \text{Pooling}\left(\boldsymbol x_p^{ \text{Top} K \cdot (1-\alpha)} + \boldsymbol x_p^{\text{Non-TopK}}\right), \\
	\boldsymbol x_p^{ \text{TopK}} &= \boldsymbol x_p^{\text{Top}^{\star}K \cdot (1-\alpha)} + \boldsymbol x_p^{\text{Top} K \cdot \alpha}. 
\end{align}
% \vspace{-5em}

\subsection{Hierarchical Fusion}
\label{sec:fusion}
% \vspace{-0.5em}

Survival analysis involves various biomarkers from localized lesions and the global perspectives of pathological images and genomics. To capture a comprehensive clinical representation, we employ a hierarchical fusion approach that integrates both low-level and high-level tokens from the integration unit, facilitating survival predictions that go beyond a single level~\cite{cmta}. For the class token, we concatenate the features and apply an MLP to extract the global representation, minimizing disruption to the information flow. In terms of local information, a direct fusion of high-dimensional medical data would significantly increase computational complexity. To tackle this, we leverage low-rank multimodal fusion (LMF)~\cite{gao2024mini,zheng2023visible}, which reduces complexity by decomposing the fusion process into modality-specific transformations and applying low-rank approximations. In LMF, the low-rank transformations, denoted as $w_p$ and $w_g$, are applied to the modality-specific embeddings, which are then combined using the Hadamard product for efficient element-wise interactions while keeping the dimensionality manageable. This method reduces the complexity of fusion from $O(n_p \times n_g)$ to $O(r \times (n_p + n_g))$, where $r$ is the rank.
The output $\tilde{\boldsymbol x}_n^{(t+1)\star}$ is obtained as follows:
% \begingroup
% \small
\begin{equation}
	{
		\begin{aligned}
			\tilde{\boldsymbol x}_n^{(t+1)\star} 
			&= \left(\boldsymbol x_p^{(t)\star} \otimes \boldsymbol x_g^{(t)\star}\right) \cdot W \\
			&= \left(\boldsymbol x_p^{(t)\star} \otimes \boldsymbol x_g^{(t)\star}\right) \cdot \left( \sum_{i=1}^r w_p^{(i)} \otimes w_g^{(i)} \right) \\
			&= \left( \boldsymbol x_p^{(t)\star} \cdot \sum_{i=1}^r w_p^{(i)} \right) \circ \left( \boldsymbol x_g^{(t)\star} \cdot  \sum_{i=1}^r w_g^{(i)}  \right) .
		\end{aligned}
	}
	\label{eq:lowrank}
\end{equation}
% \endgroup

In the context, $W$ is a weight tensor of dimensions $n_p \times n_g \times n_h$, where $n_h$ corresponds to the size of the output representation. 
The terms $w_p^{(i)}$ and $w_g^{(i)}$ are the low-rank factors for the pathological and genetic modalities respectively. The dot product ($\cdot$) computes the inner product of tensors, while $\otimes$ is the Kronecker product. The Hadamard product ($\circ$) performs element-wise multiplication across matrices.
Finally, using Eq.~(\ref{eq:fusion1}-\ref{eq:fusion3}), we fuse the local information $\tilde{\boldsymbol x}_n^{(t+1)\star}$ and the global information $\tilde{\boldsymbol x}_n^{cls\star}$ to obtain the final prediction output.

	\vspace{-0.5em}
	\section{Experiments}
	\label{sec:experiment}
	% \vspace{-0.5em}
	% We conduct extensive experiments on five TCGA datasets for comprehensive evaluations. Initially, we compare AdaMHF with several SOTA methods. Furthermore, we establish a missing-modality benchmark, testifying robustness under challenging conditions. Finally, we perform ablation studies and highlight the reduction in computational complexity.

	\subsection{Datasets and Settings}
	% \vspace{-0.5em}
	To validate the effectiveness of our model, we conduct extensive experiments on five publicly available cancer datasets from The Cancer Genome Atlas (TCGA)\footnote{https://portal.gdc.cancer.gov/}, each containing paired WSIs and genomic data with ground-truth survival outcomes. The datasets include Bladder Urothelial Carcinoma (BLCA), Breast Invasive Carcinoma (BRCA), Glioblastoma \& Lower Grade Glioma (GBMLGG), Lung Adenocarcinoma (LUAD), and Uterine Corpus Endometrial Carcinoma (UCEC). We employ 5-fold cross-validation to ensure a more reliable evaluation of the model.
	
	The evaluation metric used is the concordance index (C-index), which assesses the model's ability to predict survival times across different individuals: 
	% The C-index evaluates concordance by ranking predicted survival risks against actual outcomes, with higher values indicating better predictive accuracy. 
	% The C-index is calculated as follows:
	
	% \begingroup
	% \small
	\vspace{-0.8em}
	\begin{equation}
		\text{C-index} = \frac{1}{n(n-1)} \sum_{i=1}^{n} \sum_{j= 1}^{n} I(T_i<  T_j)(1-c_j) ,
		% \vspace{-0.6em}
	\end{equation}
	% \endgroup
	where $n$ is the number of patients, $I(\cdot)$ is the indicator function (returns 1 for truth and 0 otherwise). $T$ is the predicted survival time, and $c$ is the right censorship status, indicating whether the survival time for the individual is censored.
	
	\vspace{-0.3em}
	% \vspace{-0.8em}
	\subsection{Comparison with State-of-the-Art Methods}
% \vspace{-0.3em}
	% \vspace{-0.5em}
	We compare AdaMHF against various baseline methods. For genetic data, we employ SNN~\cite{snn} and SNNTrans~\cite{snn} as baselines, while for pathology, we utilize multi-instance learning (MIL) methods such as AttnMIL~\cite{attentionmil}, CLAM~\cite{clam}, TransMIL~\cite{transmil}, DeepAttnMISL\cite{deepmil} and DTFD-MIL\cite{zhang2022dtfd}. Additionally, we evaluate AdaMHF alongside several multimodal SOTA methods, including CMTA~\cite{cmta}, MCTA~\cite{coattention}, MOTCat~\cite{multi_transport}, HFBSurv\cite{li2022hfbsurv}, GPDBN\cite{wang2021gpdbn} and Porpoise~\cite{porpoise}. The results, presented in Table~\ref{tab:main}, highlight AdaMHF’s superior performance in one-versus-all comparisons, with particularly strong outcomes on the BRCA and BLCA datasets, showing improvements of 1.7\%, 1.8\%, and an overall improvement of 1.8\% compared to prior methods.

\begin{table}[t]
    \vspace{-1.0em}
    \centering
    \caption{The missing-modality benchmark on the BLCA, GBMLGG, LUAD, and BRCA datasets. 
Models marked with stars denote our re-implemented versions.}
    \vspace{-0.5em}
    \label{tab:miss}
    \scalebox{0.92}{
        \begin{tabular}{lccccc}
            \toprule
            \textbf{Models} & \textbf{Missing Type} & \textbf{BLCA} & \textbf{GBMLGG} & \textbf{LUAD} & \textbf{BRCA} \\ 
            \midrule
            MCAT*~\cite{coattention}         & Geno.          & 0.606            &  0.735           & 0.566            & \underline{0.614} \\ 
            CMTA*~\cite{cmta}         & Geno.          & 0.610           & 0.739           & \underline{0.598}         & 0.612 \\ 
            MOTCat*~\cite{multi_transport} & Geno. & \underline{0.612}           & \underline{0.740}            & 0.571           & 0.608 \\ 
            PORPOISE*~\cite{porpoise}  & Geno.          & 0.523            & 0.619            & 0.567            & 0.478 \\ 
            \textbf{AdaMHF (Ours)}     & Geno.          & \textbf{0.632}   & \textbf{0.754}   & \textbf{0.632}   & \textbf{0.626}\\ 
            \midrule
            MCAT*~\cite{coattention}         & Patho.         & \underline{0.660}          & 0.818            & 0.647           & 0.641 \\ 
            CMTA*~\cite{cmta}         & Patho.         & 0.655            & \underline{0.837}             & \underline{0.675}      & 0.639 \\ 
            MOTCat*~\cite{multi_transport} & Patho. & 0.641            & 0.831            & 0.659            & \underline{0.657}\\ 
            PORPOISE*~\cite{porpoise}  & Patho.         & 0.601            & 0.790            & 0.609            & 0.615 \\ 
            \textbf{AdaMHF (Ours)}     & Patho.         & \textbf{0.698}   & \textbf{0.855}   & \textbf{0.691}   & \textbf{0.672} \\ 
            \bottomrule
        \end{tabular}
    }
    \vspace{-2.3em}
\end{table}

	% \endgroup

\begin{figure*}[ht]
		\vspace{-1.0em}
		\centering
		\includegraphics[width=1.0\textwidth]{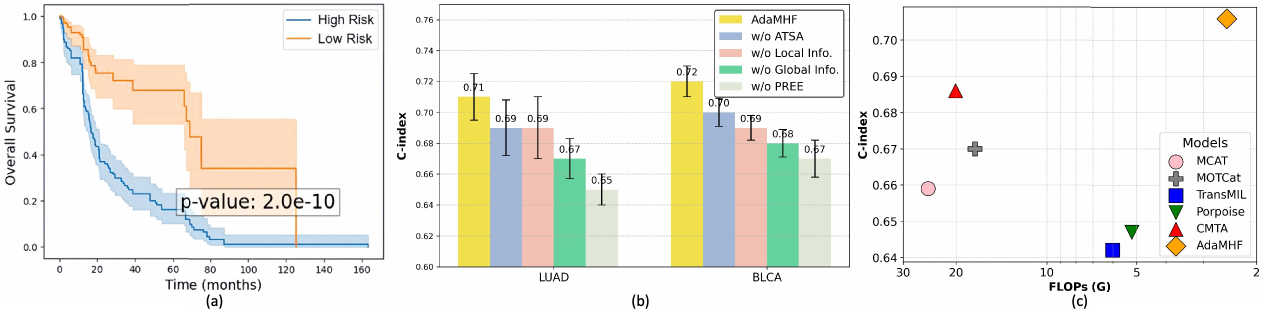}
		\vspace{-2.3em}
		\caption{Statistical analysis and ablation study: (a) Kaplan-Meier curves for statistical analysis of AdaMHF on the LUAD datasets. (b) Ablation study of AdaMHF key components over LUAD and BLCA datasets. (c) Computational complexity analysis.}
		\label{fig:km_abla}
		\vspace{-0.8em}
	\end{figure*}

	To further evaluate the efficacy of AdaMHF, we conduct a statistical analysis, with the Kaplan-Meier curves shown in Fig.~\ref{fig:km_abla}(a). Patients were divided into two groups based on the median value of the risk scores predicted by our method. We use the log-rank test to compute p-values, assessing the statistical significance of the differences between the low-risk group (blue) and the high-risk group (orange). As illustrated in Fig.~\ref{fig:km_abla}(a), the p-value is well below 0.05, indicating a significant distinction between the two groups. 
	
	% \begin{figure}[ht]
		% \centering
		% \includegraphics[width=0.50\textwidth ]{Conference-LaTeX-template_10-17-19/img/km.png}
		
		% \vspace{-1.2em}
		% \caption{Kaplan-Meier curves for statistical analysis of AdaMHF on the LUAD dataset.}
		% \label{fig:km_curves}
		% \vspace{-1.2em}
		% \end{figure}

	\vspace{-0.3em}
	\subsection{Missing-Modality Benchmark}
	% \vspace{-0.1em}
	Existing methods for survival analysis in multimodal prediction often assume the availability of both tissue images and genetic profiles. However, in clinical practice, pathological images may be missing due to the inability to perform biopsies, and genomic data may be unavailable due to blood instability or necrotic tissue. To resolve these real-world challenges, we established a missing-modality benchmark to evaluate model performance under such conditions. Specifically, we provide complete multimodal data for training. In inferencing, we simulate missing modalities by omitting data from specific sources, evaluating the model's performance on survival prediction with incomplete information.

	As shown in Table~\ref{tab:miss}, on the TCGA datasets, our model demonstrates superior performance in missing-modality scenarios. When only the genomic modality is missing, AdaMHF surpasses competitors by 3.4\% on the LUAD dataset, and when the pathological modality is missing, it exceeds by 3.8\% on the BLCA dataset.
	Notably, our model with missing modalities outperforms most models in Table~\ref{tab:main} that use full modalities, highlighting the robustness of our approach. Implementation details, overall Kaplan-Meier curves, and a comprehensive overview of the established benchmark are provided in the supplementary material.

% \begin{figure*}[ht]
% \centering
% \includegraphics[width=1.0\textwidth]{Conference-LaTeX-template_10-17-19/img/km.pdf}
% % \vspace{-2.5em}
% \caption{Kaplan-Meier Curves for Statistical Analysis of AdaMHF on the comprehensive TCGA Datasets.}
% \label{fig:km_app}
% % \vspace{-1.2em}
% \end{figure*}

	% \begingroup
	% \small
\begin{table}[t]
	\centering
	\caption{Component-wise performance comparison under various constraints on the LUAD dataset.}
	\label{tab:app_abla}
	\vspace{-0.5em}
	\scalebox{0.99}{
		\begin{tabular}{@{}llc@{}}
			\toprule
			\textbf{Module}        & \textbf{Variant}             & \textbf{C-index} \\ \midrule
			\multirow{5}{*}{PREE}  
			& w/o MLP Residual Connection            & $0.680 \pm 0.043$                   \\
			& Standard MoE                 & $0.664 \pm 0.038$                   \\ 
			& 1-Layer                      & $0.667 \pm 0.017$                   \\ 
			& 2-Layers                     & $\underline{0.689 \pm 0.021}$                   \\ 
			& 3-Layers                     & $0.706 \pm 0.024$                   \\ 
			\midrule
			ATSA                   & w/o Preserving Strategy      & $0.683 \pm 0.038$                   \\ 
			\midrule
			\multicolumn{2}{@{}l}{\textbf{AdaMHF}}                &\textbf{0.706} $\pm$ \textbf{0.024}      \\ 
			\bottomrule
		\end{tabular}}
	\vspace{-1.2em}
\end{table}

	% \endgroup

    \begin{table}[t]
    \centering
    %\vspace{-0.5em}
    \caption{Contribution of different modules to FLOPs reduction. A checkmark (\checkmark) indicates the absence of the respective module.}
    \label{tab:app_flops}
    \vspace{-0.5em}
    \begin{tabular}{c c c c c}
    \toprule
    \textbf{ID} & \textbf{w/o ATSA} & \textbf{w/o PREE} & \textbf{w/o LMF} & \textbf{FLOPs (G) ($\downarrow$)} \\
    \midrule
    1 & \checkmark &  \checkmark & \checkmark & 21.44 \\
    2 & \checkmark & & & 13.80 \\
    3 & & \checkmark & & 11.28 \\
    4 & &  & \checkmark & 7.63 \\
    AdaMHF &  & &  & 2.51 \\
    \bottomrule
    \end{tabular}
    \vspace{-1.5em}
\end{table}

	\vspace{-0.2em}
	\subsection{Ablation Study}
	
	% \vspace{-0.4em}
	\noindent \textbf{Component Ablation.} We conduct additional experiments to evaluate the effectiveness of key components in our model. Specifically, we analyze the impact of the ATSA and PREE, as well as the absence of global or local information. As shown in Fig.~\ref{fig:km_abla}(b), the results demonstrate that PREE has the most significant impact on prediction accuracy, achieving a fluctuation of 6\%, while AdaMHF effectively balances the contributions of its components, resulting in optimal performance.
	
	% \vspace{0.3em}
	\noindent \textbf{Module-Level Discussion:} To evaluate the contributions of different structures, we conduct ablation experiments. Table~\ref{tab:app_abla} shows that removing MLP residual connections decreases performance by 2.6\%. Replacing PREE with a vanilla Mixture of Experts (MoE) structure causes an additional 4.2\% drop. Similarly, eliminating the token-preservation operation in the ATSA module results in a 2.3\% reduction in the C-index due to the loss of critical information. Varying the number of PREE layers shows that increasing layers improves accuracy initially, but performance plateaus after three layers. To achieve an optimal balance between prediction accuracy and computational efficiency, we choose three PREE layers as the final configuration. For more details, refer to the supplementary.

\vspace{-0.5em}	
\subsection{Complexity Analysis}
    
	\noindent \textbf{Complexity Comparation.} Equations~\ref{eq:atsa3} and~\ref{eq:lowrank} show that both the PREE and ATSA modules and low-rank fusion significantly lower computational complexity. To validate its efficiency, we compare AdaMHF with several SOTA methods on the LUAD dataset, assessing computational complexity. Fig.~\ref{fig:km_abla}(c) shows that AdaMHF not only achieves the highest C-index but also reduces FLOPs by nearly 10 times, consistently outperforming existing SOTA methods while significantly lowering computational costs.
	
	% \vspace{0.3em}
\noindent \textbf{Impact of Each Module on FLOPs}:
 We evaluate the individual contributions of ATSA, PREE, and low-rank fusion (LMF) in AdaMHF on FLOPs reduction. ATSA is analyzed by directly dropping it and denoting its absence. For PREE, we activate all three expert layers to simulate its exclusion. LMF is replaced by the outer product operation to assess its impact. Table ~\ref{tab:app_flops} reveals that ATSA achieves an impressive reduction of 8.9 GFLOPs, while the combination of all modules collectively reduces FLOPs by 88.3\% compared to the baseline. These findings underscore the effectiveness of the proposed AdaMHF framework in enhancing computational efficiency.

\vspace{-0.1em}

	% \vspace{-0.9em}
	\section{Conclusion}
	% \vspace{-/0.6em}
	
In this paper, we propose a novel framework, adaptive multimodal hierarchical fusion for survival prediction (AdaMHF). The framework integrates a progressive residual experts expansion (PREE) module for dynamically extracting features and an adaptive token selection and aggregation (ATSA) module to optimize efficiency and minimize the impact of missing and redundant features, complemented by hierarchical fusion for integrating local and global features.  The combination of these components effectively addresses sparsity and heterogeneity in survival prediction at minimal computational cost. Extensive experiments on five TCGA datasets validate our model's superior performance compared to various SOTA methods. Additionally, results on the proposed missing-modality benchmark demonstrate the adaptability and resilience of our approach.
	%advancing survival analysis toward practical applications and providing a foundational tool for future research in this area.

\bibliographystyle{IEEEbib}
\bibliography{icme2025references}

\clearpage
\section{SUPPLEMENTARY MATERIAL}
\appendix

\section{Detailed Experimental Settings }
\subsection{Computational Resources}
\label{app:gpu}
All preprocessing steps, as well as the training and evaluation processes of the AdaMHF model, are executed on NVIDIA V100 GPUs with 32 GB of memory and A100 GPUs with 40 GB of memory. 

\subsection{Hyperparameter Configuration}
\label{app:para}

The Cancer Genome Atlas (TCGA)\footnote{\url{https://portal.gdc.cancer.gov/}} database comprises five publicly available cancer datasets: Bladder Urothelial Carcinoma (BLCA, $n = 373$), Breast Invasive Carcinoma (BRCA, $n = 956$), Glioblastoma and Lower Grade Glioma (GBMLGG, $n = 569$), Lung Adenocarcinoma (LUAD, $n = 453$), and Uterine Corpus Endometrial Carcinoma (UCEC, $n = 480$).

During the training process of AdaMHF on different datasets, the key hyperparameters are summarized in Table~\ref{tab:params}. In Table~\ref{tab:params}, $\lambda$ represents the weight assigned to the loss function, the specific definition of which is provided in Sec. ~\ref{app:loss}. The optimizer employed is the Adam optimizer, with the learning rate controlling the step size of the optimizer. Due to the high memory consumption of the clinical data, the batch size is set small to mitigate memory usage.
 The parameter $N_{s} $ refers to the number of randomly selected patches from pre-extracted features of WSIs that are used as input, which helps reduce the input dimensions and further alleviate memory constraints. The \textit{Epoch} denotes the number of training iterations. For each epoch, both training and evaluation are performed.
$\beta$ represents the ratio of global information to local information during the hierarchical fusion process. 

The entire training procedure follows a 5-fold cross-validation protocol. In this setting, the dataset is partitioned into five equal subsets (folds). During each iteration, four folds are used for training, while the remaining fold serves as the validation set. This process is repeated five times, ensuring that each fold is used as a validation set once. After completing the five iterations, the final model performance is calculated as the average of the evaluation metrics across all folds, and the standard deviation (std) is also reported to quantify the variance across the folds. This cross-validation technique provides a more robust evaluation by reducing the potential impact of overfitting to a single train-test split.
   % \vspace{-1.5em}
\begin{table}[ht]
    \centering
    \caption{Parameter table for AdaMHF across different TCGA datasets.}
    \vspace{-0.5em}
    \resizebox{0.5\textwidth}{!}{ % 将表格宽度缩放到文本宽度
        \begin{tabular}{@{\hskip 5pt}l@{\hskip 15pt}c@{\hskip 15pt}c@{\hskip 15pt}c@{\hskip 15pt}c@{\hskip 15pt}c@{\hskip 15pt}c@{\hskip 5pt}}
            \toprule
            Dataset & $\lambda$ & \textbf{lr} & \textbf{Batch Size} & $N_{s}$ & \textbf{Epoch} & $\beta$ \\ 
            \midrule
            BLCA    & 0.10   & 0.001  & 1  & 2048  & 50    & 0.05  \\
            BRCA    & 0.50   & 0.005  & 1  & 2048  & 20   & 0.05  \\
            UCEC    & 0.50  & 0.0001 & 2  & 2048  & 50    & 0.10 \\
            LUAD    & 0.10   & 0.001   & 2 & 2048  & 50   & 0.20  \\
            GBMLGG  & 1.00  & 0.001  & 1 & 2048  & 30   & 0.05 \\
            \bottomrule
        \end{tabular}
    }
    \label{tab:params}
\end{table}

\section{Additional Results on AdaMHF}

\subsection{Further Benchmark Results}
\label{app:benchmark}

As shown in Table~\ref{tab:app_miss}, we introduce a comprehensive benchmark specifically designed for survival analysis in scenarios involving missing modalities. This benchmark is built upon a more extensive and diverse dataset, enabling a broader and more robust evaluation. By simulating various real-world missing-modality conditions, this benchmark provides deeper insights into model adaptability and resilience, offering a more realistic assessment compared to existing methods. Furthermore, it establishes a standardized testing ground for future research in survival prediction, facilitating comparisons across different multimodal fusion frameworks.

\begin{table*}[ht]
\centering
\caption{The complete missing modality benchmark on five TCGA public datasets. Models marked with stars indicate those re-implemented by us. The evaluation metric remains the C-index. The best results are highlighted in \textbf{bold}, while the second-best results are \underline{underlined}.}
\vspace{-0.5em}
\label{tab:app_miss}
\small % 调整表格整体大小
\begin{tabular}{lccccccc}
\toprule
\textbf{Model} & \textbf{Missing Type} & \textbf{BLCA} & \textbf{GBMLGG} & \textbf{BRCA} & \textbf{LUAD} & \textbf{UCEC} & \textbf{Overall} \\ \midrule
CMTA*~\cite{cmta}       & Geno.            & 0.610 $\pm$ 0.023 & 0.739 $\pm$ 0.028 & \underline{0.618 $\pm$ 0.042} & \underline{0.598 $\pm$ 0.021} & 0.607 $\pm$ 0.023 & \underline{0.634}\\ 
MCAT*~\cite{coattention}         & Geno.          & 0.606 $\pm$ 0.041 & 0.735 $\pm$ 0.035 & 0.614 $\pm$ 0.040 & 0.566 $\pm$ 0.001 & \underline{0.621 $\pm$ 0.038} & 0.628\\ 
PORPOISE*~\cite{porpoise}  & Geno.          & 0.523 $\pm$ 0.001 & 0.619 $\pm$ 0.001 & 0.478 $\pm$ 0.002 & 0.567 $\pm$ 0.002 & 0.602 $\pm$ 0.005 & 0.558\\ 
MOTCat*~\cite{multi_transport} & Geno.  & \underline{0.612 $\pm$ 0.015} & \underline{0.741 $\pm$ 0.022} & 0.608 $\pm$ 0.021 & 0.571 $\pm$ 0.036 & 0.616 $\pm$ 0.036 & 0.630 \\ 
\textbf{AdaMHF (Ours)}            & Geno.            & \textbf{0.623 $\pm$ 0.022} & \textbf{0.754 $\pm$ 0.019} & \textbf{0.624 $\pm$ 0.011} & \textbf{0.632 $\pm$ 0.012} & \textbf{0.633 $\pm$ 0.011} & \textbf{0.653} \\ \midrule

CMTA*~\cite{cmta}       & Patho.           & 0.625 $\pm$ 0.037 & \underline{0.837 $\pm$ 0.021} & 0.639 $\pm$ 0.012 & \underline{0.678 $\pm$ 0.014} & 0.622 $\pm$ 0.018 & 0.680\\ 
MCAT*~\cite{coattention}         & Patho.         & \underline{0.660 $\pm$ 0.034} & 0.818 $\pm$ 0.040 & 0.641 $\pm$ 0.039 & 0.647 $\pm$ 0.027 & \underline{0.650 $\pm$ 0.042} & \underline{0.683}\\ 
PORPOISE*~\cite{porpoise}  & Patho.         & 0.601 $\pm$ 0.001 & 0.790 $\pm$ 0.013 & 0.615 $\pm$ 0.003 & 0.609 $\pm$ 0.215 & 0.555 $\pm$ 0.004 & 0.634\\ 
MOTCat*~\cite{multi_transport} & Patho. & 0.641 $\pm$ 0.022 & 0.831 $\pm$ 0.029 & \underline{0.657 $\pm$ 0.033} & 0.639 $\pm$ 0.032 & 0.642 $\pm$ 0.023 & 0.682\\ 
\textbf{AdaMHF (Ours)}           & Patho.           & \textbf{0.698 $\pm$ 0.012} & \textbf{0.855 $\pm$ 0.034} & \textbf{0.669 $\pm$ 0.038} & \textbf{0.691 $\pm$ 0.022} & \textbf{0.684 $\pm$ 0.021} & \textbf{0.719} \\ 
\bottomrule
\end{tabular}
\end{table*}

\subsection{Comprehensive Ablation Experiments}
\label{app:abla}

\begin{itemize}

    \item \textbf{Fusion Rate ($\beta$):} We investigate the impact of varying the fusion ratio $\beta$ between global and local information. As shown in Fig.~\ref{fig:app_abla}, achieving the highest C-index requires balancing holistic and fragmented information. As outlined in the main text, identifying the optimal fusion rate is crucial for achieving peak performance. The best rate is achieved at $\beta = 0.2$, likely because local information tends to be more abundant, thus a lower proportion for local information is both appropriate and intuitive.

    \item \textbf{$K$ and $\alpha$ in ATSA:} In the ATSA module, the parameters $K$ and $\alpha$ are determined via a routing mechanism. For this analysis, we manually set these hyperparameters to evaluate their effects, as depicted in Fig.~\ref{fig:app_abla}. When one of $K$ or $\alpha$ is fixed, the other is dynamically obtained, and the specified hyperparameter applies simultaneously to both the pathology modality and gene modality. The findings indicate that dynamically assigning values for $K$ and $\alpha$ enhances the model's adaptability and flexibility.

%     \item \textbf{The Impact of Each Module on FLOPs}:  
%     We investigate the contribution of ATSA and PREE in AdaMHF to the reduction of FLOPs. For ATSA, we use a direct masking approach, denoted as w/o ATSA. For PREE, we compare it with the scenario where all three layers of experts are activated, denoted as w/o PREE. The detailed results can be found in Table~\ref{tab:app_flops}, where it is evident that both PREE and ATSA reduce computational overhead to varying degrees, and their combination yields even more significant improvements.

% \begin{table}[ht!]
% \centering
% \caption{Contribution of different modules to FLOPs reduction on the LUAD dataset}
% \label{tab:app_flops}
% \vspace{1em}
% \begin{tabular}{c c  | c}
% \toprule
% \textbf{ATSA} &\textbf{ PREE} & \textbf{Flops (G)}   \\
% \midrule
% \ding{55} & \ding{55} & 18.44  \\
% \ding{55} &\checkmark  & 14.80  \\
% \checkmark & \ding{55} & 7.63  \\
% \checkmark & \checkmark & 2.51  \\
% \bottomrule
% \end{tabular}
% \end{table}

\begin{figure}[ht]
\centering
% \vspace{1.5em}
\includegraphics[width=0.5\textwidth]{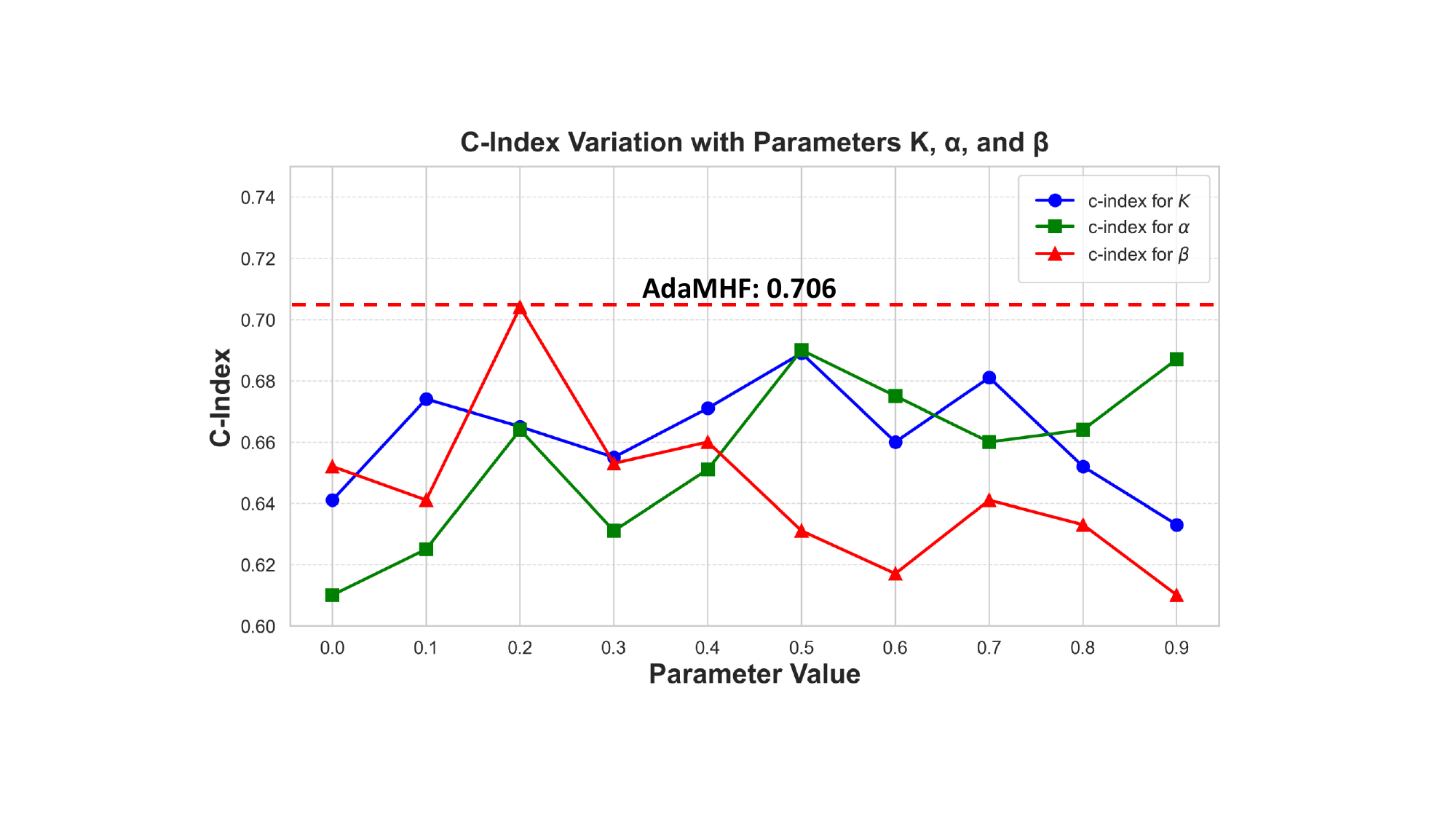}
\vspace{-0.5em}
\caption{
Ablation study on the effects of parameters \textcolor[rgb]{0,0.5,0}{$\alpha$}, \textcolor{red}{$\beta$}, and \textcolor{blue}{$K$} on model performance.
}
\label{fig:app_abla} % 放在 \caption 之后
\end{figure}

\end{itemize}

\subsection{Additional Kaplan-Meier Curves}
\label{app:km}

\begin{figure*}[ht]
\centering
\includegraphics[width=1.0\textwidth]{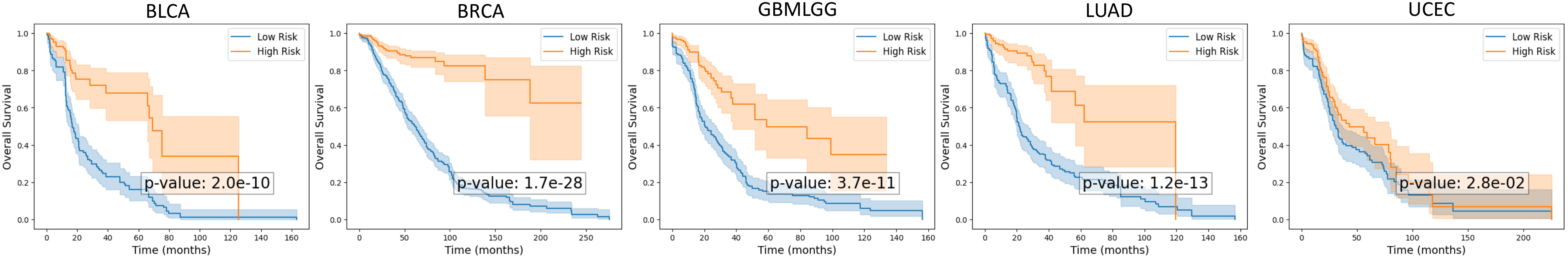}
\vspace{-0.9em}
\caption{Kaplan-Meier curves for statistical analysis of AdaMHF on the comprehensive TCGA datasets.}
\label{fig:km_app}
% \vspace{-1.2em}
\end{figure*}

In this section, we present additional Kaplan-Meier curves to demonstrate the survival probabilities associated with the AdaMHF framework across various datasets. As shown in Figure~\ref{fig:km_app}, these curves provide a visual representation of survival outcomes over time, enabling a clear comparison of different subgroups within the data. This expanded analysis highlights the adaptability of AdaMHF to varying data distributions, offering deeper insights into its performance in subgroup-specific contexts. Furthermore, these curves facilitate a better understanding of how the model captures survival trends, establishing a comprehensive foundation for evaluating its effectiveness in clinical applications.

\section{Details of Loss Functions and Model Architecture}

\subsection{Detailed Loss Functions}
\label{app:loss}

In addition to the loss function \(L_{\text{surv}}\), we introduce an 
auxiliary loss \(L_{\text{align}}\) to ensure that the extracted features from the WSIs and omics data for the same patient exhibit similarity and consistency. Specifically, \(L_{\text{align}}\) constrains the representations $\boldsymbol x_p^{cls \star} $ and $\boldsymbol x_g^{cls\star} $ using the L1 norm to facilitate the learning of important shared information. The overall loss can thus be expressed as:
\begin{equation}
    L=L_{surv}+\lambda L_{align}.
\end{equation}

\subsection{Preprocessing Network for Input Clinical Data}
\label{app:clam}
For the preprocessing of whole slide images (WSIs), we adhered to the methodology outlined in ~\cite{attentionmil,cmta,coattention}, utilizing the code provided by CLAM\footnote{\url{https://github.com/mahmoodlab/CLAM/tree/master}}. Specifically, CLAM loads and wraps a pre-trained Vision Transformer (ViT) model (vit\_large\_patch16\_224), removes the output layer, and employs it as a headless feature extractor. This process ultimately yields $n_p$ patch features, denoted as $\boldsymbol x_p $. In contrast, the preprocessing of genomic data is seamlessly integrated into the overall training workflow of AdaMHF, employing a self-normalizing neural network (SNN) for processing rather than handling it separately. The omic data is categorized into six groups: 1) Tumor Suppression, 2) Oncogenesis, 3) Protein Kinases, 4) Cellular Differentiation, 5) Transcription, and 6) Cytokines and Growth. Subsequently, these six groups are concatenated and input into the SNN to obtain $\boldsymbol x_g $.

\subsection{Experts Types in PREE}

\subsubsection{Rationale for Expert Selection in PREE}
\label{app:experts}
For the experts in the PREE module, we select modality-specific experts to adaptively extract features according to the unique characteristics of each modality. For pathological images, we employ the Convolutional Neural Network (CNN) for feature extraction. CNNs are particularly effective for pathological images due to their capacity to capture spatial hierarchies and local patterns, such as texture, morphology, and structural details, which are crucial in identifying abnormalities in histopathological slides. The local receptive fields and weight-sharing mechanisms in CNNs allow them to efficiently learn from the high-resolution, gigapixel whole slide images (WSIs), making them ideal for capturing intricate tissue structures that are indicative of disease progression.
For genomic data, we utilize the same network architecture as used in the preprocessing stage—the Self-Normalizing Neural Network (SNN). SNNs are highly appropriate for genomic data because they maintain stable activation distributions during training, even in deep architectures. This is particularly beneficial when dealing with genomic data, which often involves high-dimensional, sparse, and noisy features. The self-normalizing properties of SNNs help to mitigate issues such as vanishing and exploding gradients, ensuring more reliable and consistent learning of complex relationships between gene expressions and patient outcomes. This makes them highly efficient in capturing the nonlinear interactions present in genomic pathways.

\subsubsection{Convolutional Neural Network}
\label{cnn}
Convolutional Neural Networks (CNNs) are a class of deep neural networks particularly effective for analyzing visual data, making them a cornerstone in pathological image-based survival prediction. In this context, CNNs process whole slide images (WSIs) to automatically and adaptively learn spatial hierarchies of features through backpropagation. Their convolutional layers are designed to capture essential pathological features, such as lesion regions, cellular structures, and their boundaries, which are critical for accurate survival prediction. These layers are followed by pooling layers that reduce data dimensionality, improving computational efficiency while retaining key diagnostic information. By leveraging their hierarchical feature extraction capabilities, CNNs play a pivotal role in identifying predictive patterns within complex and high-resolution pathological slides.

\subsubsection{Self-Normalizing Neural Network}
\label{snn}

Self-Normalizing Neural Networks (SNNs) have emerged as a powerful tool in genomic survival prediction due to their ability to maintain stable activation distributions during training. This stability is achieved through the use of scaled exponential linear units (SELUs), an activation function uniquely designed to promote a zero mean and unit variance across network layers. Such self-normalization reduces the reliance on additional techniques like batch normalization, facilitating faster convergence and mitigating the risks of vanishing or exploding gradients.

In the context of survival prediction using genomic data, SNNs excel in handling the high-dimensional and heterogeneous nature of gene expression profiles. By ensuring stable activations, they effectively capture subtle patterns and variations in genomic features that are predictive of survival outcomes. This property is particularly advantageous when dealing with complex gene interactions, allowing SNNs to robustly identify prognostic biomarkers critical for accurate predictions.

\subsection{Parameters of Fixed MLP in PREE}
\label{app:MLP}

In the PREE module, we employ a fixed MLP for residual connections to enhance the extraction of shallow features, which are crucial for providing essential morphological and structural details necessary for identifying early indicators of disease. This approach ensures stability in both training and results. 
Given that genomic data and pathological images share similar low-level information, both contributing to the structured analysis of lesions, we adopt identical parameter initialization for the fixed MLP in both PREE-G and PREE-P modules.

For the initialization of the MLP parameters, we utilize a pre-trained model provided by CLAM, specifically a modified pre-trained ViT (vit\_large\_patch16\_224) from which the classifier head has been removed. We extract the MLP parameters from the first layer of the encoder to serve as the parameters for the fixed MLP, thereby stabilizing the training process.

\section{Underlying Network Structure of the AdaMHF Model}
\subsection{Mixture of Experts}
\label{moe}

The Mixture of Experts (MoE) architecture offers a robust foundation for addressing the unique challenges of survival prediction in multimodal settings, such as those involving pathological images and genomic data. By leveraging multiple specialized neural networks, MoE is well-suited to extract complementary features from these heterogeneous data types.

The gating mechanism, a central component of the MoE framework, dynamically routes data to the most appropriate expert based on intra-modality attributes, such as spatial features in pathological images or sequence patterns in genomic data. This dynamic routing not only ensures efficient computational resource allocation but also enhances the adaptability of the model to missing or incomplete modalities—a common issue in real-world survival prediction tasks.

Compared to traditional models that uniformly process multimodal inputs, MoE conditional computation activates only the experts relevant to each modality, reducing computational overhead while maintaining high prediction accuracy. Moreover, the modular design of MoE mitigates the risk of overfitting by diversifying the learning process, enabling the model to handle intra-modal heterogeneity (e.g., variations in tumor morphology or genomic mutations) and inter-modal disparities (e.g., differing biological characteristics of pathological and genomic data). This makes MoE an effective framework for survival prediction based on the fusion of pathological images and genomic data, aligning with the goals of AdaMHF in addressing sparsity, heterogeneity, and missing modality scenarios.

% \vspace{-em}
\subsection{Cross Attention}
\label{crossattention}
The Cross Attention mechanism is a fundamental architectural component within the AdaMHF framework, designed to facilitate effective interaction between genomic data and pathological images for survival prediction. By enabling dynamic interplay between genomic modality and pathological modality, the Cross Attention mechanism enhances the model's ability to capture complementary information, improving its ability to address inter-modal heterogeneity and increase prediction accuracy.

This mechanism allows the model's \textit{query} component to selectively emphasize relevant features across modalities by interacting with distinct \textit{key} and \textit{value} components from each input. Specifically, it computes compatibility scores between query-key pairs derived from genomic ($\boldsymbol{x}_g$) and pathological ($\boldsymbol{x}_p$) representations. These scores guide the weighting of value vectors, enabling the model to synthesize information in a contextually enriched manner. This process ensures that modality-specific features are effectively aligned and integrated, ultimately enhancing the overall performance of the AdaMHF framework. In AdaMHF, the Cross-Modal Attention module operates as follows:

\begin{align}
    V &= W_p^1 \boldsymbol{x}_p, \quad U = W_g^1 \boldsymbol{x}_g, \\[10pt]
    X &= V^\top U, \\[10pt]
    P &= \text{ColumnSoftmax}(X), \quad G = \text{RowSoftmax}(X), \\[10pt]
    \boldsymbol{x}_p^{\star} &= P^\top \boldsymbol{x}_p \, W_p^2, \quad 
    \boldsymbol{x}_g^{\star} = G \, \boldsymbol{x}_g \, W_g^2.
\end{align}

Here, $\boldsymbol{x}_p$ and $\boldsymbol{x}_g$ represent the pathological and genomic feature embeddings respectively. Trainable parameters $W_p^1$, $W_g^1$, $W_p^2$, and $W_g^2$ project these embeddings into latent spaces suitable for interaction. The raw attention scores in matrix $X$ are computed as the dot product of the projected embeddings, and they are normalized into attention matrices $P$ (column-wise) and $G$ (row-wise) using the Softmax function. These attention matrices facilitate the generation of refined outputs $\boldsymbol{x}_p^{\star}$ and $\boldsymbol{x}_g^{\star}$, highlighting the most relevant cross-modal features.

By integrating cross attention, AdaMHF aligns genomic and pathological representations, capturing nuanced interactions between these modalities. This enhances the framework's ability to overcome challenges such as sparsity and heterogeneity, contributing to improved survival prediction performance.

	% Supplement
% \vspace{-0.4em}

\end{document}